\documentclass[10pt,journal,compsoc]{IEEEtran}

\ifCLASSOPTIONcompsoc
\else
  \usepackage[noadjust]{cite}
\fi
\ifCLASSINFOpdf
  \usepackage[pdftex]{graphicx}
\else
\fi

\usepackage{algorithm,algorithmic}

\usepackage[none]{hyphenat}

\begin{document}
%
\title{Pre-Sorted Tsetlin Machine\\ (The Genetic K-Medoid Method)}
%
%
%
%

\author{Jordan~Morris,~\IEEEmembership{Member,~IEEE}, Alex Yakovlev,~\IEEEmembership{Fellow,~IEEE}
\IEEEcompsocitemizethanks{\IEEEcompsocthanksitem Jordan Morris and Alex Yakovlev are with the $\mu$Systems design group, Newcastle University, Newcastle, NE1 7RU .\protect\\
E-mail: \{jordan.morris, alex.yakovlev\}@newcastle.ac.uk}

}

\IEEEtitleabstractindextext{%
\begin{abstract}
This paper proposes a machine learning pre-sort stage to traditional supervised learning using Tsetlin Machines. Initially, K data-points are identified from the dataset using an expedited genetic algorithm to solve the maximum dispersion problem. These are then used as the initial placement to run the K-Medoid clustering algorithm. Finally, an expedited genetic algorithm is used to align K independent Tsetlin Machines by maximising hamming distance. For MNIST level classification problems, results demonstrate up to 10\% improvement in accuracy, $\mathtt{\sim}$383X reduction in training time and $\mathtt{\sim}$99X reduction in inference time.
\end{abstract}

}

\maketitle

\IEEEdisplaynontitleabstractindextext

%
\IEEEpeerreviewmaketitle

\ifCLASSOPTIONcompsoc
\IEEEraisesectionheading{\section{Introduction}\label{sec:introduction}}
\else
\section{Introduction}
\label{sec:introduction}
\fi

%
%
%
%
\IEEEPARstart{T}{he} recent surge in neural network-based machine learning and artificial intelligence has propelled many in the field to revisit more traditional learning methodologies. Many of these offer simpler feedback mechanisms and simpler primitives than the perceptron. This often translates into simpler hardware and faster software. \\
\indent One architecture to gain significant traction is the Tsetlin Machine \cite{GRANMO}, which is based on propositional logic and uses Tsetlin Automata to identify common sub-patterns in binarized data from a given entropy. Instead of backpropagation using stochastic gradient decent to optimize biases and weights, Tsetlin Machine employs a game-theoretic feedback mechanism that optimises for Nash equilibria. \\
\indent Whilst the Tsetlin Machine is a capable algorithm, much of the computation time is spent focusing on outliers within the data \cite{RAFIEV}, for which, more computationally-efficient handling exists. This paper therefore investigates how the training times, inference times and accuracy of the Tsetlin Machine are impacted when datasets are pre-sorted. Specifically, the contributions of this paper are:-

\begin{itemize}
  \item Using an expedited genetic algorithm to "conceptually arrange" the binarized datasets and identify K datapoints that provide the maximum dispersion based on hamming distance.
  \item Using the K-Medoid clustering algorithm to sort same-class datapoints into K clusters based on hamming distance.
  \item Using an expedited genetic algorithm to align the newly-formed clusters into K independent Tsetlin Machines.
  \item An evaluation of how this pre-sorting methodology impacts the accuracy, training time and inference time for MNIST level classification problems.
  \item A discussion on how the outlined stages may be optimised in hardware.
\end{itemize}

\indent This paper consists of 7 sections. Section 2 provides a brief overview of the Tsetlin Machine. Section 3 details the proposed pre-sort architecture. Section 4 delineates the experiment methodologies. Section 5 presents the results. Section 6 discusses how such improvments in accuracy and performance are achieved. Section 7 concludes the paper.

\section{Tsetlin Machine}

An exhaustive description of the Tsetlin Machine, complete with formal mathematical derivation of the game theoretic feedback mechanism, may be found in the seminal paper \cite{GRANMO}. To aid the reader, a brief description of the key features of the algorithm are provided here.

\subsection{Tsetlin Automata}

\begin{figure}[h!]
\begin{center}
	\includegraphics[scale=0.52]{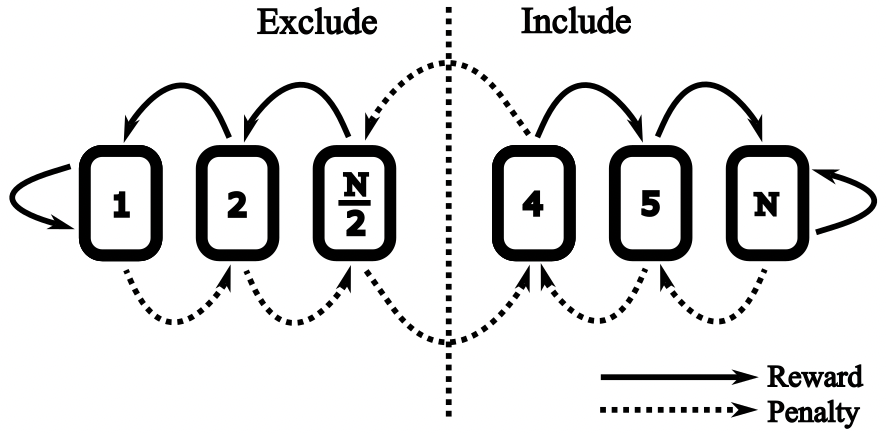}
\end{center}
\caption{Tsetlin Automaton}
\label{fig:automaton}
\end{figure}

The Tsetlin Automaton provides the elementary primitive for the Tsetlin Machine algorithm. In it's most basic form, the Tsetlin Automaton is simply a linear state machine consisting of N states, upon which, actions may be taken to change the current state. An example of a Tsetlin Automata with N=6 may be seen in Figure \ref{fig:automaton}.

\clearpage

The N states are split into two equal regions, "Exclude" and "Include". These represent whether or not a particular bit should be included in a given sub-pattern. Feedback to the automaton may be provided in the form of two actions: Reward, Penalty. A reward action indicates that the feedback mechanism suspects that the current state of the automaton falls within the correct include/exclude region. The reward action therefore reinforces the current decision by pushing the current state of the automaton further away from the include/exclude boundary. A penalty action indicates that the feedback mechanism suspects that the current state of the automaton falls within the incorrect region. The penalty action therefore weakens the current decision by pushing the current state of the automaton towards the include/exclude boundary. If the state was already on the include/exclude boundary when the penalty action was received, the boundary is traversed and the decision on whether the particular bit represented by the automaton is included or excluded is changed accordingly.

\subsection{Clauses}

\begin{figure}[h!]
\begin{center}
	\includegraphics[scale=0.47]{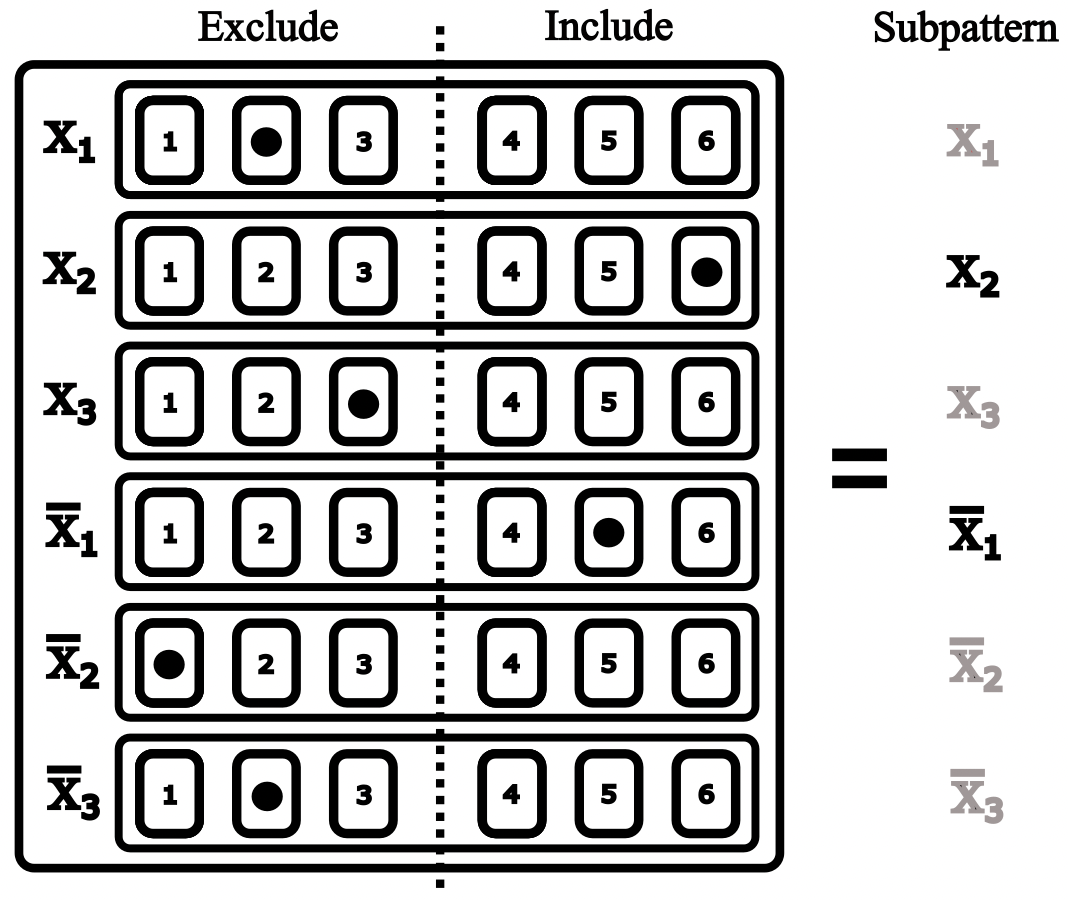}
\end{center}
\caption{Tsetlin Machine Clause}
\label{fig:clause}
\end{figure}

One Tsetlin Automaton is used to represent each bit in the binarized input feature (F). An additional Tsetlin Automaton is used to represent the negated value of each bit in the binarized input feature. These are commonly referred to as literals (L). Hence L = 2F. A clause with F=3 (L=6) may be seen in Figure \ref{fig:clause}.

The figure shows two automata with include decisions and four automata with exclude decisions. The subpattern created by this clause is $\{X_{2},\overline{X}_{1}\}$. Only the two included bits of the input feature shall be checked (both $X_{3}$ and $\overline{X}_{3}$ are excluded). The clause shall become active if the first bit in the feature is zero ($\overline{X}_{1}$) and the second bit in the feature is one ($X_{2}$). For all other combinations of the first two bits of the input feature, the clause shall become inactive.

\subsection{Tsetlin Machine Architecture}

The basic structure of a Tsetlin Machine is dependent on the number of classes required for the dataset. For two classes, a simple mono structure is sufficient. For three or more classes, a slightly more sophisticated multi-class structure is required.

\subsubsection{Mono Tsetlin Machine Architecture}

\begin{figure}[h!]
\begin{center}
	\includegraphics[scale=0.49]{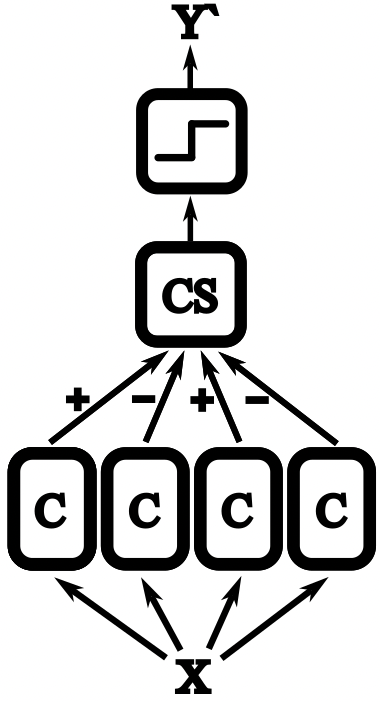}
\end{center}
\caption{Mono Tsetlin Machine Architecture}
\label{fig:mono}
\end{figure}

A high level depiction of the mono Tsetlin Machine architecture may be seen in Figure \ref{fig:mono}.

Clauses (C) are instantiated as "clause-pairs", with one assigned a positive polarity and one assigned a negative polarity. Assuming a dataset of two classes (Class 0 and Class 1), the negative clauses may be assigned to Class 0 and the positive clauses to Class 1. A datapoint is distributed to all clauses in a single action. If a clause is active, it votes with its polarity (+1/-1). If it is inactive, it does not vote (and therefore has no impact on further calculations). These positive and negative votes are accumulated into a class sum (CS). A simple threshold is then applied to the class sum. If the class sum is below the threshold, the predicted class is Class 0. If the class sum is above the threshold, the predicted class is Class 1.

\begin{figure}[h!]
\begin{center}
	\includegraphics[scale=0.38]{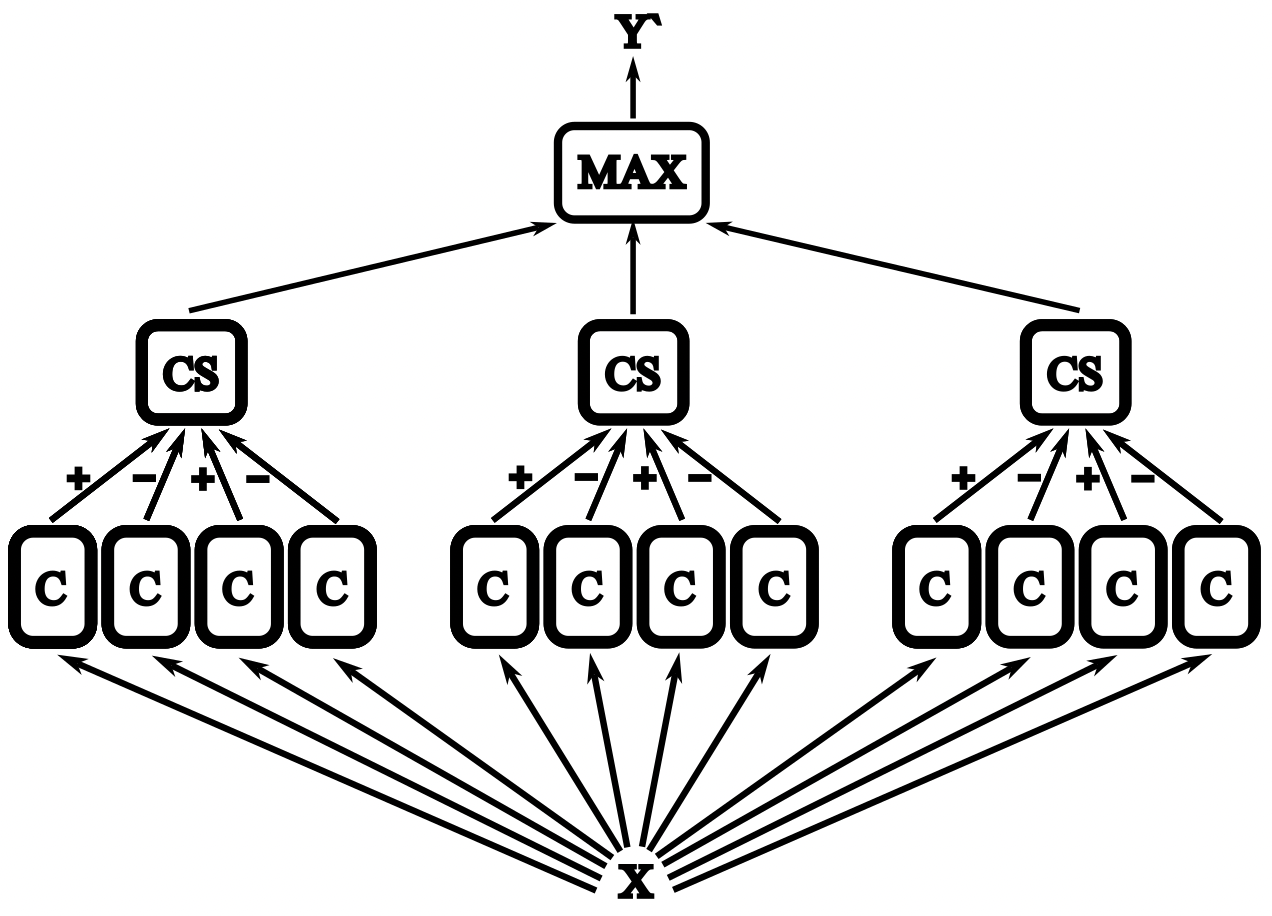}
\end{center}
\caption{Multi-Class Tsetlin Machine Architecture}
\label{fig:multi}
\end{figure}

\subsubsection{Multi-Class Tsetlin Machine Architecture}

For three or more classes, the mono Tsetlin Machine architecture is repeated, one for each class, and the thresholding is replaced with an argmax. A high level depiction of a 3-class Tsetlin Machine may be seen in Figure \ref{fig:multi}.

In this instance, the binarized input feature is distributed to all clauses, the three class sums are calculated and the largest class sum becomes the predicted class.

\section{Pre-Sort Architecture}

\begin{figure}[h!]
\begin{center}
	\includegraphics[scale=0.25]{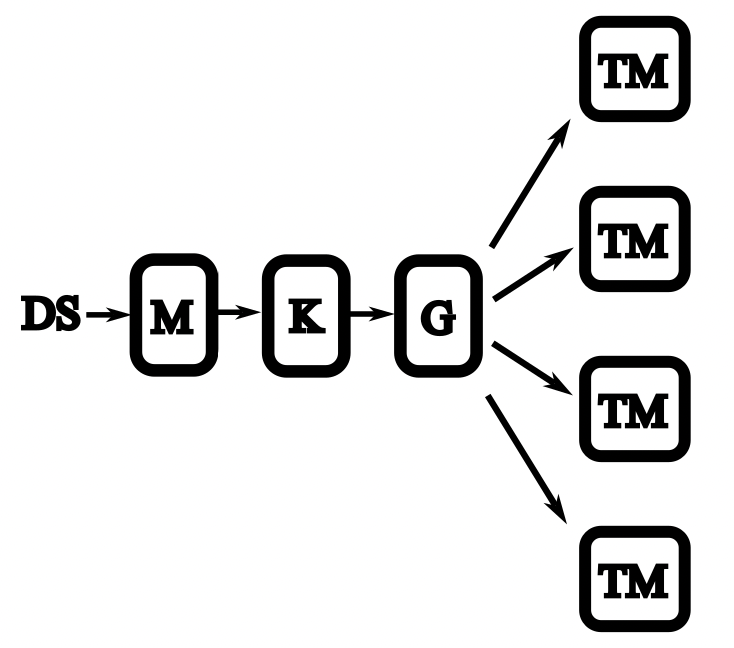}
\end{center}
\caption{Pre-Sort Architecture}
\label{fig:presort}
\end{figure}

The proposed pre-sort architecture can be split into three distinct stages: Binary Maximum Dispersion (M), Binary K-Medoid (K) and Genetic Class Alignment (G). These stages are run sequentially on the dataset. The result is a rapid sorting of data that may be split into K data subsets that are then distributed to smaller Tsetlin Machines. A high level overview of the pre-sort architecture may be seen in Figure \ref{fig:presort}.

\subsection{Binary Maximum Dispersion}

\begin{figure}[h!]
\begin{center}
	\includegraphics[scale=0.6]{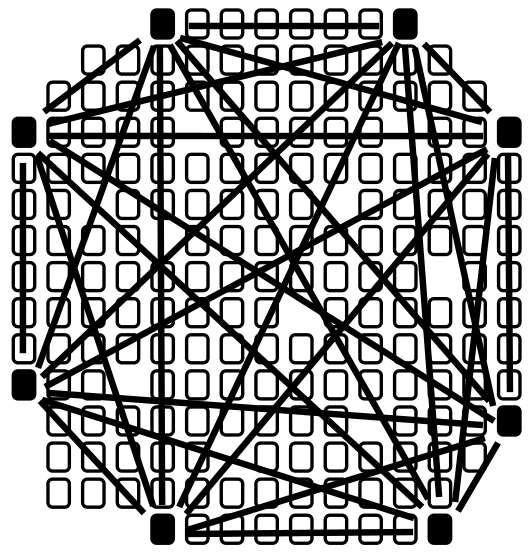}
\end{center}
\caption{Binary Maximum Dispersion}
\label{fig:diff}
\end{figure}

The purpose of this stage is to identify K datapoints within each class of the dataset that represent the maximum cumulative difference possible. This is commonly referred to as solving the maximum dispersion problem. For binarized datasets, there is usually no ordinality (one handwritten digit isn't inherently greater or lesser than another). However, a "distance" between two binarized datapoints can be measured using the hamming distance (the number of bits different between them). In this manner, the K maximally-distant datapoints may be identified and the remaining data "conceptually arranged" by hamming distance between them. A visualiation of this process for K=8 datapoints may be seen in Figure \ref{fig:diff}.

To find the datapoints, an expedited genetic algorithm is used to determine the greatest cumulative hamming distance between K chosen datapoints. The final K datapoints need only be a rough estimate at this stage as they are updated by the next stage. The genetic algorithm therefore does not require execution until exhaustion and can be parallelized to further reduce run time.

\subsection{Binary K-Medoid Algorithm}

\begin{figure}[h!]
\begin{center}
	\includegraphics[scale=0.6]{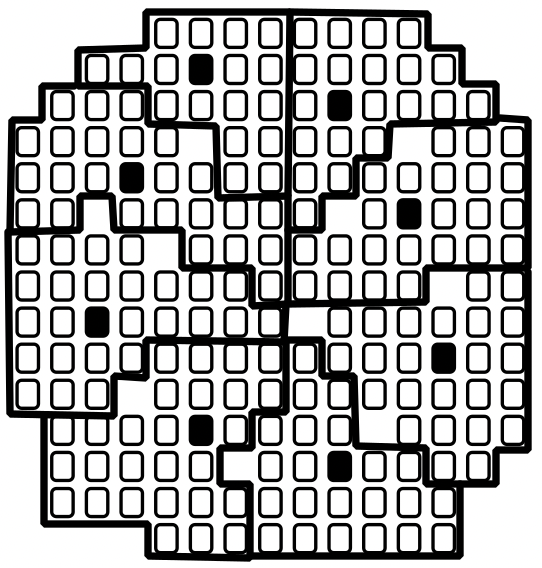}
\end{center}
\caption{Binary K-Medoid Clustering}
\label{fig:kmed}
\end{figure}

The Binary K-Medoid Algorithm performs the bulk of the work in the pre-sort architecture. Unlike the more common K-Means clustering algorithm, K-Medoid chooses an actual datapoint to cluster on. This makes the algorithm more robust when handling datasets with considerable outliers. Moreover, given that most binarized data has no ordinality, a calculated "mean" would not provide any meaningful insight. \\
\indent The algorithm first takes the K datapoints from the Binary Maximum Dispersion stage. These become the initial medoids. All non-medoid datapoints are then assigned to their nearest medoid based on hamming distance to form K clusters. Next, the cumulative distance from each datapoint is calculated to all other datapoints within the same cluster. The datapoint in each cluster with the smallest cumulative distance is then selected to be the new medoid for that cluster. All datapoints are then assigned to their nearest medoid and the process is iterated. The algorithm converges when the medoids are unchanged between iteration cycles. \\
\indent A visualisation of the final result of this process may be seen in Figure \ref{fig:kmed}.

\subsection{Genetic Class Alignment}

\begin{figure}[h!]
\begin{center}
	\includegraphics[scale=0.6]{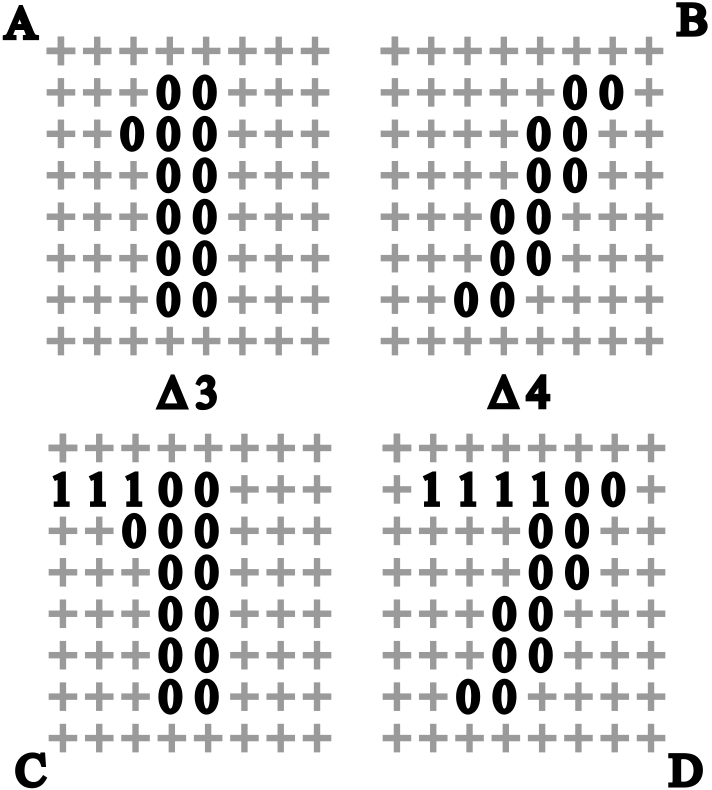}
\end{center}
\caption{Pre-Alignment Class Medoids}
\label{fig:med-init}
\end{figure}

\begin{figure}[h!]
\begin{center}
	\includegraphics[scale=0.6]{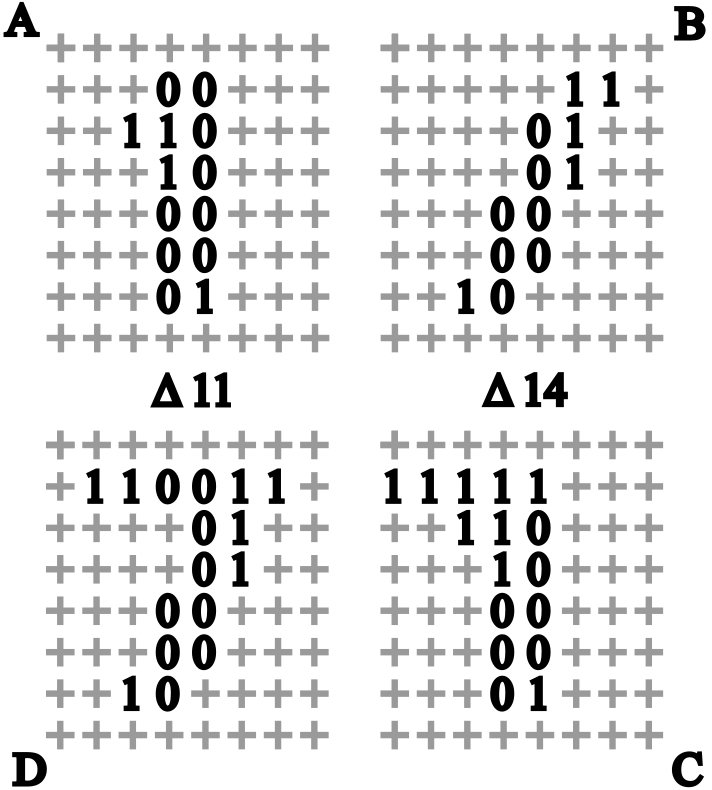}
\end{center}
\caption{Post-Alignment Class Medoids}
\label{fig:med-sort}
\end{figure}

The dataset now consists of K clusters within each class. Given that the Tsetlin Machine functions by mining frequent sub-patterns and using these to differentiate between the classes, the optimal strategy is to align clusters based on maximal inter-class hamming distance and to distribute these to K separate Tsetlin Machines. \\
\indent To illustrate this concept, Figure \ref{fig:med-init} shows 4 final medoids from the previous stage for a handwritten digit dataset. Two are from Class "1" and two are from Class "7". Medoid A shows a classic number 1 with a vertical stroke whereas Medoid B shows an italicised style 1 with a rightward slanting stroke. Medoid C shows a 7 with a vertical line and leftward stroke, whereas Medoid D shows a more classic rightward slanting stroke and leftward stroke. The "+" symbol indicates that no marking was detected for this bit in the datapoint. The "0" symbol indicates a marking was detected for this bit and this marking location is shared with the medoid vertically aligned with it. The "1" symbol indicates a marking was detected for this bit and this marking location is not shared with the medoid vertically aligned with it. Using these symbols to aid us, it is clear that there are only 3 bits different (hamming distance) between Medoid A and Medoid C and 4 bits different between Medoid B and Medoid D. In fact, Medoid A is wholly encompassed within Medoid C and Medoid B is wholly encompassed within Medoid D. This makes classification by sub-pattern identification relatively difficult between these medoids. \\
\indent Figure \ref{fig:med-sort} shows a different alignment with Medoids C and D switched. The hamming distances between the vertically aligned medoids have now significantly increased. This leads to more unique sub-patterns that can be learned to differentiate between Medoids A and D and Medoids B and C, making the task of the Testlin Machine easier.
\indent To implement this concept, the final medoids from the K-Medoid stage (K medoids within each class) are placed with their class peers into 1D horizontal arrays. These arrays are then vertically stacked as shown in Figures \ref{fig:med-init} and \ref{fig:med-sort} (such that each row represents one class). The total cumulative hamming distance between each medoid in a vertical stack is calculated and then summed for all vertical stacks to give a total hamming distance for the 2D configuration. An expedited genetic algorithm is used to search for the 2D configuration with the largest total hamming distance. The columns (containing one cluster from each class) are then used to separate the dataset in K smaller datasets. These are then processed by their own independent Tsetlin Machine.

\section{Experiments}

The pre-sort architecture was written in C++ to match the Tsetlin Machine code provided on the main CAIR (Centre for Artificial Intelligence Research) repository \cite{CAIR}. Four MNIST style datasets were chosen for classification. The original MNIST dataset, containing handwritten digits from 0 to 9, EMNIST, the Extended MNIST dataset, KMNIST, the Kuzushiji MNIST datatset consisting of handwritten Japanese characters (contains class-overlap, multiple characters per class) and FMNIST, the Fashion MNIST dataset consisting of 28x28 graphical representations of common clothing items. These datasets were binarized using a simple thresholding method \cite{ABEYRATHNA} (($<X$ set to 0), ($>=X$ set to 1)). \\
\indent Datasets were classified with a varying number of medoids (2,4,8,16,32). Two runtime strategies were chosen. The first consisted of running both the baseline Tsetlin Machine algorithm and proposed pre-sort method for 100 epochs, then running inference. This determined the training time, inference time and accuracy. However, as the number of medoids was increased during the experiment, it became apparent that some pre-sort Tsetlin Machines reached 100\% accuracy within 4-6 epochs, after which, it was superfluous to continue training. The second strategy was therefore to run an inference between each training cycle for these pre-sort Tsetlin Machines. When 100\% accuracy was reached, the training was concluded. The training runtime for this method includes the inference time required to check the accuracy after each training epoch. The results presented show the fastest of these two methods for the pre-sort training time, as both are legitimate training strategies. All dataset/medoid points were performed 10 times using unique 80/20 train/test splits of the datasets and the average (mean) is presented. 4000 clauses per class were used for the baseline Testlin Machine implementation as suggested on the CAIR repository. For the pre-sorted Tsetlin Machines, 4000 clauses per class were used in total across all distributed Tsetlin Machines (eg K=2, $\mathtt{\sim}$2000 clauses per class per machine). As the clustering is non-uniform, the number of clauses assigned to each distributed Tsetlin Machine was apportioned based on the total number of datapoints assigned to that Tsetlin Machine. Other options are available, but were not explored. All runs were performed single-threaded on a first-gen M1 Macbook Pro.

\section{Results}

\begin{figure}[h!]
\begin{center}
	\includegraphics[scale=0.5]{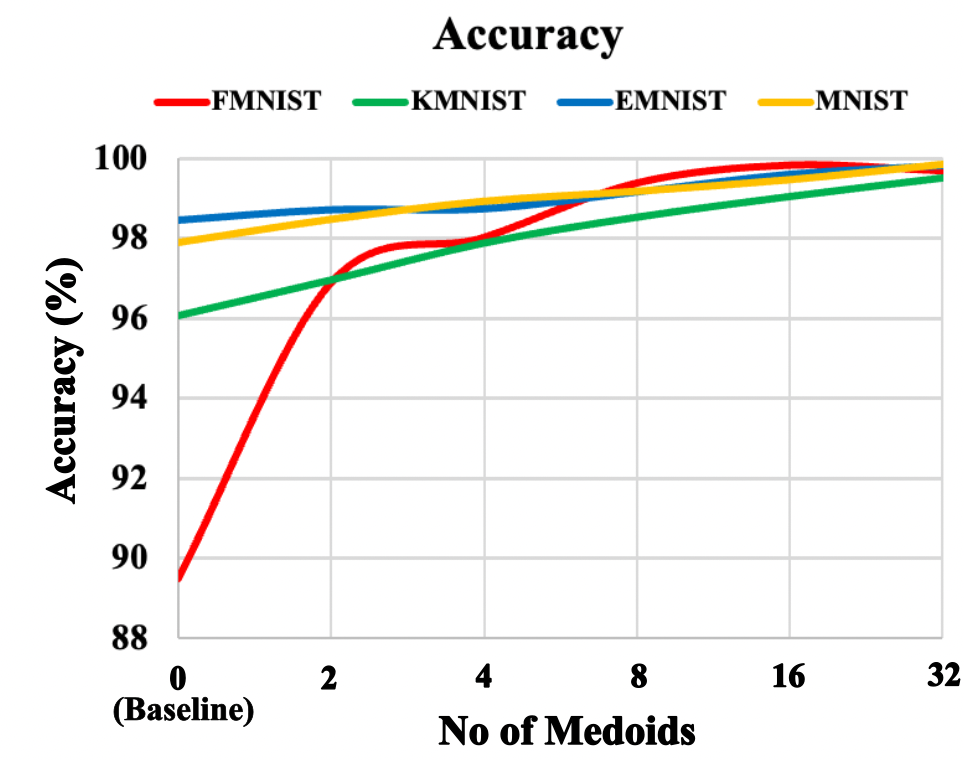}
\end{center}
\caption{Accuracy}
\label{fig:graph-acc}
\end{figure}

\begin{figure}[h!]
\begin{center}
	\includegraphics[scale=0.5]{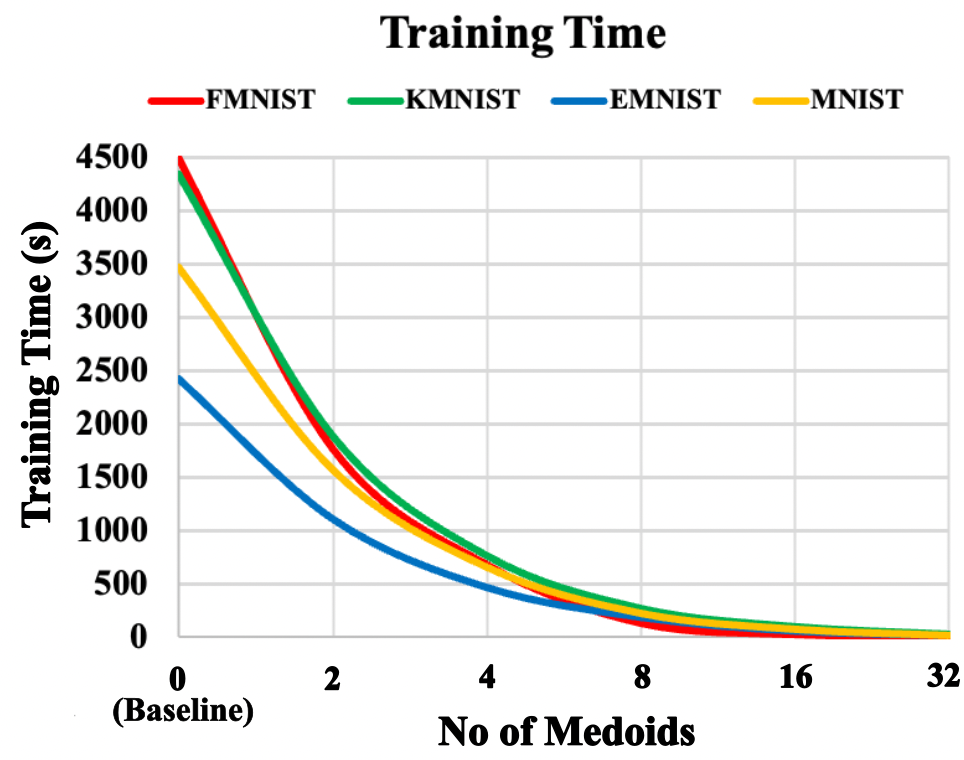}
\end{center}
\caption{Training Time}
\label{fig:graph-train}
\end{figure}

\begin{figure}[h!]
\begin{center}
	\includegraphics[scale=0.5]{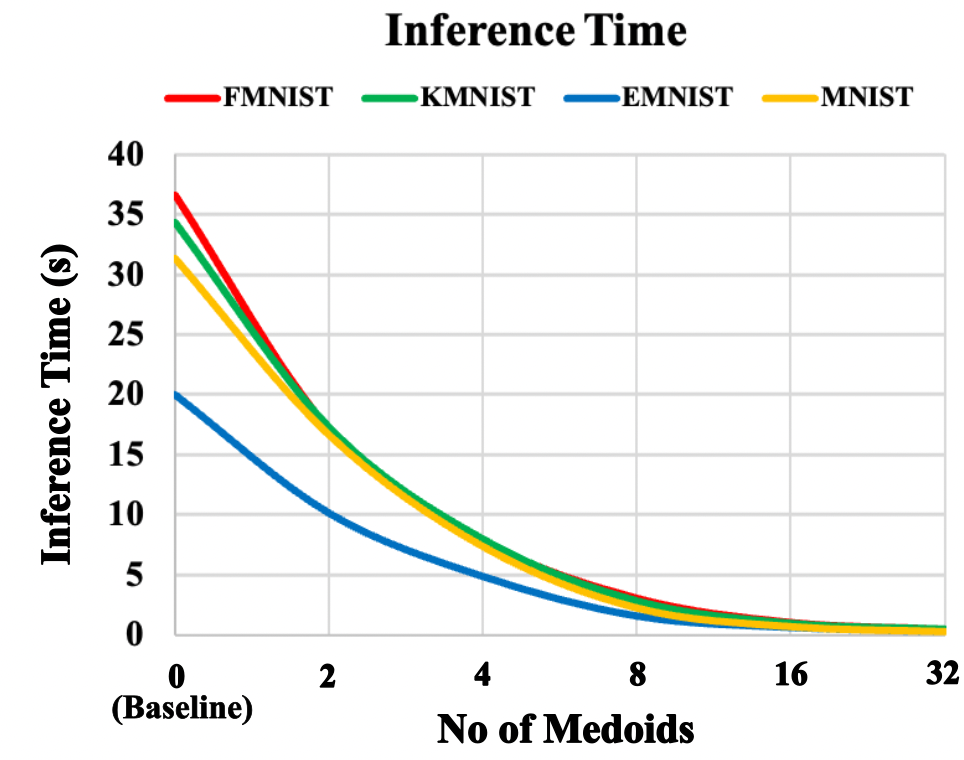}
\end{center}
\caption{Inference Time}
\label{fig:graph-test}
\end{figure}

Figure \ref{fig:graph-acc} shows the classification test accuracy between the baseline Tsetlin Machine algorithm through varying medoid configurations for the four chosen datasets. Critical datapoints for the test accuracy can be found in Table 1. \\
\indent Figure \ref{fig:graph-train} shows the training time between the baseline Tsetlin Machine algorithm through varying medoid configurations for the four chosen datasets. Critical datapoints for the training time can be found in Table 2. \\
\indent Figure \ref{fig:graph-test} shows the inference time between the baseline Tsetlin Machine algorithm through varying medoid configurations for the four chosen datasets. Critical datapoints for the inference time can be found in Table 3.

\begin{table}[h!]
\centering
\caption{Accuracy}
\begin{tabular}{lcccc}
\hline
\textbf{}       & \textbf{FMNIST} & \textbf{KMNIST} & \textbf{EMNIST} & \textbf{MNIST} \\ \hline
Baseline        &     89.50\%     &     96.07\%     &      98.45\%    &    97.88\%     \\
Pre-Sorted      &     99.81\%     &     99.50\%     &      99.86\%    &    99.86\%     \\ \hline
\textbf{Gain}   &     10.32\%     &     3.43\%      &      1.41\%     &    1.98\%      \\ \hline
\end{tabular}
\end{table}

\begin{table}[h!]
\centering
\caption{Training Time}
\begin{tabular}{lcccc}
\hline
\textbf{}       & \textbf{FMNIST} & \textbf{KMNIST} & \textbf{EMNIST} & \textbf{MNIST} \\ \hline
Baseline        &     4484 s      &      4336 s     &     2424 s      &    3472 s      \\
Pre-Sorted      &     11.72 s     &      30.89 s    &     15.58 s     &    15.58 s      \\ \hline
\textbf{Speed Up}  &     382.5X   &      140.4X     &     155.6X      &    222.9X       \\ \hline
\end{tabular}
\end{table}

\begin{table}[h!]
\centering
\caption{Inference Time}
\begin{tabular}{lcccc}
\hline
\textbf{}       & \textbf{FMNIST} & \textbf{KMNIST} & \textbf{EMNIST} & \textbf{MNIST} \\ \hline
Baseline        &     36.6 s      &      34.3 s     &     20.0 s      &    31.4 s      \\
Pre-Sorted      &     0.43 s      &      0.43 s     &     0.32 s      &    0.32 s      \\ \hline
\textbf{Speed Up}   &    85.8X    &       79.7X     &     62.9X       &    98.8X       \\ \hline
\end{tabular}
\end{table}

\section{Discussion}

It can be seen from Figure \ref{fig:graph-acc} that for all datasets, the test accuracy improves for the pre-sorted Tsetlin Machine when compared to the baseline implementation (in some instances, markedly so). There are several reasons for this. One reason is class-overlap (more than one "symbol" or "character" per class). The results demonstrate that the baseline Testlin Machine has difficulty with this concept as it mines frequent sub-patterns during training to form the clauses. With class-overlap, the sub-patterns are different for each unique symbol and the feedback mechanism can alter clauses from one "symbol" with feedback from another "symbol" within the same class. In pre-sorted Tsetlin Machine, dissimilar "symbols" are separated and farmed out to independent smaller Tsetlin Machines, so this situation is inherently handled. \\
\indent Another issue is translation (identical patterns occurring in differing locations between datapoints). In the baseline Tsetlin Machine, different sub-patterns have to be learned for each location that a sub-pattern appears in. A sliding window mechanism has been tested to solve this issue with baseline Tsetlin Machine \cite{GRANMO-CONV}, however, this adds considerable complexity and dramatically increases the training and inference times. In pre-sorted Tsetlin Machine, this is inherently handled as the different translations are simply sorted into separate clusters and processed in independent Tsetlin Machines. \\
\indent It can be seen from Figure \ref{fig:graph-train} that for all datasets, the training time is drastically reduced for the pre-sorted Tsetlin Machine with respect to the baseline implementation. There are several reasons for this. Principally, fewer clauses are run per training cycle. This is because in the baseline Tsetlin Machine, 4000 clauses are updated for every datapoint passed into a class. In pre-sorted Tsetlin Machine, only 1/K * 4000 clauses are updated on average per datapoint. This immediately results in a training runtime reduction by a factor of K. Another reason is that having pre-simplified the problem, the test accuracy reached 100\% for some pre-sorted Tsetlin Machines after only a few epochs. It was therefore faster to adopt a strategy of testing for this condition and then terminating training rather than running for an extensive period of time. \\
\indent It can be seen from Figure \ref{fig:graph-test} that for all datasets, the inference time is drastically reduced for the pre-sorted Tsetlin Machine with respect to the baseline implementation. The reason for this is identical to the reduction in training time. The number of clauses required for an inference is reduced by a factor of K, lowering the computational complexity and therefore reducing the runtime. \\
\indent As the entire pre-sort architecture may be parallelised, a further train/inference run time reduction of at least a factor of K is achievable against the results herein presented simply by running multi-threaded.

\subsection{Momentary Mastery}

As the original Tsetlin Machine is split out into smaller independent Tsetlin Machines in the proposed pre-sort scheme, a unique opportunity is presented for problems that only require "one-shot" training and inference (eg training is performed, inference is performed, the results are stored and the model deleted). As each datapoint only requires one smaller independent Tsetlin Machine for training and inference and the datapoints are grouped together, only one of the smaller independent Tsetlin Machines needs to exist at any given time. This would result in a memory footprint reduction by a factor of K, provided that a single threaded training/inference speed was acceptable. This is a considerable saving for embedded/edge devices.

\subsection{Hardware Impact}

The heart of the Tsetlin Machine algorithm is really three simple instructions: AND, XNOR and popcount. The AND and XNOR operations are used to quickly match set/unset bits between the clauses and the datapoint whilst calculating the clause outputs. Popcount is used to quickly count set bits in words corresponding to clause votes during training and inference. Optimising hardware for these instructions can improve the performance of the algorithm. \\
\indent Whilst Section 3 provided sound conceptual reasoning behind why each of the three stages of the pre-sort architecture were chosen, perhaps more importantly, there is synergy with the hardware requirements of Tsetlin Machine. The heart of all three pre-sort stages is the hamming distance calculation: XOR followed by popcount. Whilst AND, XOR (bitwise inequality) and XNOR (bitwise equality) are common logical operators, popcount is typically omitted in RISC architectures. Hence, extending these architectures with the additional popcount instruction improves the entire proposed pre-sort architecture, not just the Tsetlin Machine.

\section{Final Comments}
This work proposed and investigated a pre-sort architecture for Tsetlin Machines. Results were presented that demonstrated up to 10\% improvement in accuracy, $\mathtt{\sim}$383X reduction in training time and $\mathtt{\sim}$99X reduction in inference time. A full delineation of how this was achieved was discussed, along with methods to reduce memory footprint and hardware considerations for improving performance.

\ifCLASSOPTIONcaptionsoff
  \newpage
\fi

\bibliographystyle{unsrt}
\bibliography{bare_adv.bib}

\vspace{-1.2cm}


\end{document}